\documentclass[twocolumn]{article}
\usepackage{hyperref}
\usepackage{times}
\usepackage{latexsym}

\usepackage{url}

\usepackage{booktabs}       
\usepackage{amsfonts}       
\usepackage{nicefrac}       

\usepackage{amsmath, amssymb, amsthm}
\usepackage{multirow}
\usepackage{algorithmic}
\usepackage{algorithm}
\usepackage{graphicx}
\usepackage{natbib}

\usepackage{enumitem}
\usepackage{todonotes}

\title{Neural Embedding Allocation:\\ Distributed Representations of Topic Models}

\author{
	Kamrun Naher Keya \\
	Department of Information Systems \\
	University of Maryland Baltimore County \\
	kkeya1@umbc.edu
	\and
	Yannis Papanikolaou \\
	Atypon Systems\\
	ypapanikolaou@atypon.com \\
	\and
	James R. Foulds \\
	Department of Information Systems \\
    University of Maryland Baltimore County\\
	jfoulds@umbc.edu
}

\date{}

\begin{document}
\maketitle
\begin{abstract}

Word embedding models such as the skip-gram learn vector representations of words' semantic relationships, and document embedding models learn similar representations for documents.
On the other hand, topic models provide latent representations of the documents' topical themes.
To get the benefits of these representations simultaneously, we propose a unifying algorithm, called \emph{neural embedding allocation} (NEA), which deconstructs topic models into interpretable vector-space embeddings of words, topics, documents, authors, and so on, by learning neural embeddings to mimic the topic models. We showcase NEA's effectiveness and generality on LDA, author-topic models and the recently proposed mixed membership skip gram topic model and achieve better performance with the embeddings compared to several state-of-the-art models. Furthermore, we demonstrate that using NEA to smooth out the topics improves coherence scores over the original topic models when the number of topics is large. 

\end{abstract}

\section{Introduction}

In recent years, methods for automatically learning representations of text data have become an essential part of the natural language processing pipeline.
Word embedding models such as the skip-gram improve the performance of Natural Language Processing (NLP) methods by revealing the latent structural relationship between words ~\citep{mikolov2013efficient,mikolov2013distributed}. 
These embeddings have proven valuable for a variety of NLP tasks such as statistical machine translation~\cite{vaswani2013decoding}, part-of-speech tagging, chunking, and named entity recognition ~\cite{collobert2011natural}. Since word vectors encode 
distributional information, the similarity relationships between the semantic meanings of the words are reflected in the similarity of the vectors~\cite{sahlgren2008distributional}.  Extensions to document embeddings have subsequently been proposed \cite{le2014distributed}.

On the other hand, topic models such as latent Dirichlet allocation (LDA)~\cite{blei2003latent} 
construct latent representations of topical themes and of documents, and these can be used to subsequently derive representations for words~\citep{griffiths2007topics}.
Like word embeddings, topic models exploit conditional discrete distribution over words to represent high-dimensional data into a low-dimensional subspace. However, topic models do not directly capture nuanced relationships between words using vector-space embeddings, which are often important for performance on downstream NLP tasks \cite{maas2011learning}. 

We therefore desire a unified method which gains the benefits of both word embeddings (encoding nuanced semantic relationships) and topic models (recovering interpretable topical themes).  Some recent research has aimed to combine aspects of topic models and word embeddings. 
The Gaussian LDA model~\cite{das2015gaussian} tries to improve the performance of topic modeling by prioritizing the semantic information encoded in word embeddings, however, it does not aim to jointly perform word embedding. Unlike Gaussian LDA, the topical word embedding model~\cite{liu2015topical} uses LDA topic assignments of words as input to improve the resultant word embedding. In another approach, 
mixed membership word embeddings~\cite{foulds2018mixed} aim to recover domain-specific interpretable word embeddings without big data, based on topic embeddings.  

\begin{table*}[t]
\centering
\resizebox{1\textwidth}{!}{
\begin{tabular}{ccc}
\toprule
& Embedding Models &  \hspace{-5cm} Topic Models\\
\toprule
& Skip-gram & \hspace{-4cm} Naive Bayes skip-gram topic model (SGTM)\\
\midrule
\begin{minipage}[t]{3.1cm}
\ \\
\ \\
Words$|$Input Word
\end{minipage} &
\hspace{1.5cm}
\begin{minipage}[t]{10cm}
\begin{itemize}
\item	For each word in the corpus $w_i$
\begin{itemize}
    \item Draw input word $w_i \sim p_{data}(w_i)$ 
    \item For each word $w_c \in context(i)$
    \begin{itemize}
        \item Draw $w_c | w_i \propto exp({v'_{w_c}}^\intercal v_{w_i})$
    \end{itemize}
\end{itemize}
\end{itemize}
\end{minipage}
&
\hspace{-2cm}
\begin{minipage}[t]{10cm}
\begin{itemize}
\item	For each word in the corpus $w_i$
\begin{itemize}
    \item Draw input word $w_i \sim p_{data}(w_i)$ 
    \item For each word $w_c \in context(i)$
    \begin{itemize}
        \item Draw $w_c | w_i  \sim \mbox{Discrete}(\phi^{(w_i)})$
    \end{itemize}
\end{itemize}
\end{itemize}
\end{minipage}
\ \\
\ \\
\midrule
& Neural embedding allocation & \hspace{-4cm} Latent Dirichlet allocation\\
\midrule
\hspace{-0.5cm}
\begin{minipage}[t]{3.1cm}
\ \\
\ \\
Words$|$Topics
\end{minipage}
 &
\hspace{1.5cm}
\begin{minipage}[t]{10cm}
\begin{itemize}
\item	For each document  $d$
\begin{itemize}
    \item For each word in the document $w_{di}$
    \begin{itemize}
        \item Draw $z_{di}|d \sim \mbox{Discrete}(\theta^{(d)})$
        \item Draw $w_{di}|z_{di} \propto \exp({v'_{w_{di}}}^\intercal \bar{v}_{z_{di}})$
    \end{itemize}
\end{itemize}
\end{itemize}
\end{minipage}
&
\hspace{-2cm}
\begin{minipage}[t]{10cm}
\begin{itemize}
\item	For each document  $d$
\begin{itemize}
    \item For each word in the document $w_{di}$
    \begin{itemize}
        \item Draw $z_{di}|d \sim \mbox{Discrete}(\theta^{(d)})$
        \item Draw $w_{di}|z_{di}  \sim \mbox{Discrete}(\phi^{(z_{di})})$
    \end{itemize}
\end{itemize}
\end{itemize}
\end{minipage}
\ \\
\bottomrule
\end{tabular}
}
\caption{``Generative'' models of the skip-gram (top-left) and its analogous supervised naive Bayes topic model (top-right), and the \emph{`neural embedding allocation} reparameterization of the LDA topic model (bottom).\label{tab:reparameterizeTopicModel}}
\vspace{-0.5cm}
\end{table*}

In contrast, in this paper we develop a method which recovers vector-space embeddings of words, documents, topics, authors, and so on, 
based on a given topic model.  Our approach, which we call \emph{neural embedding allocation} (NEA), is to \emph{deconstruct topic models} by reparameterizing them using vector-space embeddings.  
We can view our method as \emph{learning to mimic a topic model with a skip-gram style embedding model} to reveal underlying semantic representations.  Our approach is thus reminiscent of model distillation for supervised models \cite{bucilua2006model, hinton2015distilling}.

We train NEA by minimizing the KL-divergence to the data distribution of the corresponding topic model, using a stream of simulated data from the model.
The resulting embeddings allow us to (1) improve the coherence of topic models by ``smoothing out'' noisy topics, (2) improve classification performance by producing topic-informed document vectors, and (3) construct embeddings and smoothed distributions over general topic modeling variables such as authors. 
We show the benefits and generality of our method by applying it to LDA, author-topic models (ATM)~\cite{rosen2004author}, and the recently proposed mixed membership skip gram topic model (MMSGTM)~\cite{foulds2018mixed}.

\section{Connections Between Word Embeddings and Topic Models} 
In this section, we first develop a bridge to connect word embeddings methods such as the skip-gram with topic models.  The skip-gram~\citep{mikolov2013distributed} and LDA~\cite{blei2003latent} models are summarized in Table \ref{tab:reparameterizeTopicModel} (top-left, bottom-right), where we have interpreted the skip-gram, which is discriminative, as a ``conditionally generative'' model. According to the distributional hypothesis, the skip-gram's conditional distributions over context words should be informative of the semantics of the words \cite{sahlgren2008distributional}.  
Similarly, \cite{griffiths2007topics} proposed to model semantic relationships between words under the LDA topic model, based on the predictive probability of one word given another, which they successfully used to solve a word association task.  This suggests that topic models implicitly encode semantic relationships between words, even though they are not parameterized as such, motivating methods to recover this information, as we shall propose here.  

The relationship between the skip-gram and topic models goes beyond their common ability to recover semantic representations of words.  In particular, the skip-gram and LDA both model conditional discrete distributions over words; conditioned on an input word in the former, and conditioned on a topic in the latter.  To relate the two models, we hence reinterpret the skip-gram's conditional distributions over words as ``topics'' $\phi^{(w_i)}$,  and the input words $w_i$ as observed cluster assignments, analogous to topic assignments $z$. Table~\ref{tab:reparameterizeTopicModel} (top) shows how the skip-gram can thus be re-interpreted as a certain parameterization of a fully supervised naive Bayes topic model, which \cite{foulds2018mixed} refer to as the \emph{(naive Bayes) skip-gram topic model (SGTM)}. 
A naive Bayes assumption is used in these models, as the context words are conditionally independent given their input words and the model parameters.

To understand how learning algorithms for the skip-gram are related to the SGTM, we introduce a variational interpretation of skip-gram training. It is well known that maximizing the log likelihood for a model is equivalent to minimizing the KL-divergence to the model's empirical data distribution, cf.  \citet{hinton2002training}.  
When trained via maximum likelihood estimation (MLE), the skip-gram (SG) and its corresponding topic model both aim to approximate this same empirical data distribution.  The skip-gram topic model (SGTM) can encode any set of conditional discrete distributions, and so its MLE recovers this distribution exactly.  Thus, we can see that the skip-gram, trained via MLE, also aims to approximate the MLE skip-gram topic model in a variational sense.  

More formally, consider the joint distributions $p(w_c, w_i)$ obtained by augmenting the skip-gram SG and its topic model SGTM with the empirical input word distribution $p(w_i) =p_{data}(w_i)$: $p_{SG}(w_c, w_i;\mathbf{v}, \mathbf{v}') = p(w_c|w_i; \mathbf{v}, \mathbf{v}')p_{data}(w_i)$ and $p_{SGTM}(w_c, w_i;\mathbf{\Phi}) =p(w_c|w_i; \mathbf{\Phi})p_{data}(w_i)$. 
It can readily be seen that
\begin{align*}
&D_{KL}(p_{data}(w_c|w_i)p_{data}(w_i)||p_{SG}(w_c, w_i;\mathbf{v}, \mathbf{v}'))\\
&= -\sum_{w_c, w_i} \frac{N_{w_c,w_i}}{N_{w_i}}\frac{N_{w_i}}{N} \log p(w_c|w_i;\mathbf{v}, \mathbf{v}') + \mbox{const }\\
&= -\sum_{w_c, w_i} \frac{N_{w_c,w_i}}{N}\log p(w_c|w_i;\mathbf{v}, \mathbf{v}') + \mbox{const .}
\end{align*}
By a similar argument, we also obtain  $D_{KL}(p_{data}(w_c|w_i)||p_{SGTM}(w_c, w_i;\mathbf{\Phi})) = -\sum_{w_c, w_i} \frac{N_{w_c,w_i}}{N}\log p(w_c|w_i;\mathbf{\Phi}) + \mbox{const}$.  Since the topic model's discrete distributions are unconstrained, this is minimized to zero at
\begin{align}
\hat{\phi}^{(w_i)}_{w_c} = \frac{N_{w_c,w_i}}{N_{w_i}} = p_{data}(w_c|w_i) \mbox{ .} \label{eqn:NBSGTM_MLE}
\end{align}
So maximizing the conditional log-likelihood for the skip-gram minimizes the KL-divergence to $p_{data}(w_c|w_i)p_{data}(w_i) = p_{SGTM}(w_c, w_i;\hat{\mathbf{\Phi}})$, where $\hat{\mathbf{\Phi}}$ is the MLE of the skip-gram topic model.
Therefore, \emph{the skip-gram is attempting to mimic the ``optimal'' skip-gram topic model}, by solving a variational inference problem which aims to make its distribution over input/output word pairs as similar as possible to that of the SGTM's MLE. With sufficiently high-dimensional vectors, e.g. $V \geq W$, it will be able to solve this problem exactly, assuming that a global optimum can be found.
While the above holds for maximum likelihood training, noise contrastive estimation (NCE)~\cite{gutmann2010noise,gutmann2012noise} approximates maximum likelihood estimation, and negative sampling~\cite{mikolov2013distributed} approximates NCE.  We can therefore view both of these training procedures as approximately solving the same variational problem, with some bias in their solutions due to the approximations that they make to maximum likelihood estimation.

We can also see from Equation \ref{eqn:NBSGTM_MLE} that the SGTM and SG's MLEs can be completely computed using the input/output word co-occurrence count matrix as  sufficient statistics. The skip-gram then has a global objective function that can be defined in terms of the word co-occurrence matrix, and the development of the GloVe model~\cite{pennington2014glove} as an alternative with a global objective function seems unnecessary in hindsight. \citet{levy2014neural}'s results further illustrate this point, as they find global matrix factorization objectives that are implicitly optimized by negative sampling and NCE as well. 

\section{Neural Embedding Allocation}

We have seen that the skip-gram minimizes the KL-divergence to the distribution over data at the maximum likelihood estimate of its corresponding topic model.  We can view this as \emph{learning to mimic a topic model with an embedding model}. 
The skip-gram has essentially deconstructed its topic model into nuanced vector representations which aim to encode the same information as the topic model. We therefore propose to apply this same approach, deconstructing topic models into neural embedding models, to other topic models.  

The resulting method, which we refer to as \emph{neural embedding allocation} (NEA), corresponds to reparameterizing the discrete distributions in topic models with embeddings.  The neural embedding model generally loses some model capacity relative to the topic model, but it provides \emph{vector representations} which encode valuable similarity information between words.  Following the skip-gram, by sharing the vectors between distributions, the vectors are encouraged to encode similarity relationships, as mediated by the discrete distributions and their relationships to each other.  NEA's reconstruction of the discrete distributions also \emph{smooths out noisy estimates}, leveraging the vectors' similarity patterns.


For example, we show the ``generative'' model for NEA in Table~\ref{tab:reparameterizeTopicModel} (bottom-left),  which reparameterizes the LDA model 
by topic vectors $\bar{v}_k$ and ``output'' word vectors $v'_w$ which mimic LDA's topic distributions over words, $\phi^{(k)}$, by re-encoding them using log-bilinear models.
In the generative model, $\theta^{(d)}$ draws a topic for a document and the topic vectors $\bar{v}_k$ are used as the input vectors to draw a word $v'_w$.
We can also consider a model variant where $\theta^{(d)}$ is reparameterized using a log-bilinear model, however we obtained better performance by constructing  document vectors based on topic vectors, as discussed below.

\begin{algorithm}[t]
\caption{Training NEA for LDA}\label{alg-main}
\small
\textbf{Input:} $W$ = \#Words, $K$ = \# Topics, $D$ = \# Documents,\\ $M$= Mini-batch size, trained LDA model $\boldsymbol{\Theta}_{LDA}$, $\boldsymbol{\Phi}_{LDA}$, ${Z}$

\textbf{Output:} $\boldsymbol{\Phi}_{NEA}$ = encoded $\boldsymbol{\Phi}_{LDA}$, $\boldsymbol{V}^{(W)\prime}$ = word-embeddings, $\boldsymbol{\bar{V}}^{(K)}$ = topic-embeddings, $\boldsymbol{V}^{(D)}$ = document-embeddings


\emph{\textbf{Embeddings steps:}}
\begin{itemize}
\item For each iteration $t$: \ \ \ \ //in practice, use mini-batches
	\begin{itemize}
				\item Draw a document, $d \sim unif(D)$
				\item Draw a topic, $ z \sim \boldsymbol{\Theta}_{LDA}^{(d)}$
				\item Draw a word, $ w \sim \boldsymbol{\Phi}_{LDA}^{(z_{d})}$
		\item Update [$\bar{v}_{z}$, $v_{w}'$]:= $\mbox {NEG}(\mbox{in} = z, \mbox{out} = w)$  
	\end{itemize}
\item For each document $d$ in $D$:  
    \begin{itemize}
		\item For each token $i$ in $d$:
		    \begin{itemize}
		        \item Update $v_{d}:= v_{d} + \frac{\bar{v}_{z_{di}}}{|\bar{v}_{z_{di}}|}$ 
		    \end{itemize}
		 \item Normalize $v_{d} := \frac{v_{d}}{|v_{d}|}$
	\end{itemize}
\end{itemize}
\emph{\textbf{Smoothing steps:}} Calculate $\boldsymbol{\Phi}_{NEA} \propto exp(\textbf{V}^{(W)\prime\intercal}\boldsymbol{\bar{V}}^{(K)})$
\end{algorithm}

\begin{algorithm}[t]
\caption{NEA for General Topic Models}\label{alg-general-nea}
\small
\textbf{Input:} Trained topic model of the form $P(a_0)\prod_{i=1}^n P(a_i|\mbox{parent}(a_i))$, where the \textbf{$a_i$} are discrete variables such as documents, authors, topics, words.\\
\textbf{Output:} Embeddings for each variable $\mathbf{V}^{(i)}$, $\mathbf{V}^{(i)\prime}$, smoothed distributions $P_{NEA}(a_i|\mbox{parent}(a_i))$

\emph{\textbf{Embeddings steps:}}
\begin{itemize}
\item	For each iteration  $t$:  \ \ \ \ //in practice, use mini-batches
\begin{itemize}
    \item sample ${a}_{0} \sim P({a}_{0})$
    \item For each random variable $a_i \in \{a_1 \ldots a_n\}$:
    \begin{itemize}
        \item sample ${a}_{i} \sim P(a_i|\mbox{parent}(a_i))$
        \item update $[{v}^{(i)}_{\mbox{parent}(a_i)}, {v}^{(i)\prime}_{a_{i}}]$ \\ := NEG(in=$\mbox{parent}(a_i)$, out=${a}_{i}$)
    \end{itemize}
\end{itemize}
\end{itemize}
\emph{\textbf{Smoothing steps:}}
\begin{itemize}
    \item For each random variable $a_i \in \{a_1 \ldots a_n\}$:
    \begin{itemize}
        \item $P_{NEA}(a_i|\mbox{parent}(a_i))\propto exp(v^{(i)\prime \intercal}_{a_i} v^{(i)}_{\mbox{parent}(a_i)})$
    \end{itemize}
\end{itemize}
\end{algorithm}


\begin{figure*}[t]
\centering
\resizebox{.9\textwidth}{!}{
\begin{tabular}{|c|c|c|c|c|c|c|c|}
\hline
LDA                                                                                                                                                            & NEA                                                                                                                                                               & LDA                                                                                                                                                             & NEA                                                                                                                                              & LDA                                                                                                                                                        & NEA                                                                                                                                                   & LDA                                                                                                                                            & NEA                                                                                                                                              \\ \hline
\multirow{10}{*}{\begin{tabular}[c]{@{}c@{}}corresponds\\ change\\ cut\\ exact\\ coincides\\ duplicates\\ volatility\\ trapping\\ reading\\ ters\end{tabular}} & \multirow{10}{*}{\begin{tabular}[c]{@{}c@{}}parameters\\ important\\ neural\\ change\\ results\\ report\\ cut\\ multiple\\ experiments\\ minimizing\end{tabular}} & \multirow{10}{*}{\begin{tabular}[c]{@{}c@{}}symbolics\\ addressing\\ choice\\ perturbing\\ radii\\ centered\\ damping\\ merits\\ vax\\ unexplored\end{tabular}} & \multirow{10}{*}{\begin{tabular}[c]{@{}c@{}}values\\ case\\ increase\\ systems\\ rate\\ point\\ feedback\\ input\\ reduces\\ stage\end{tabular}} & \multirow{10}{*}{\begin{tabular}[c]{@{}c@{}}ryan\\ learning\\ bit\\ inhibited\\ nice\\ automatica\\ tucson\\ infinitely\\ stacked\\ exceeded\end{tabular}} & \multirow{10}{*}{\begin{tabular}[c]{@{}c@{}}learning\\ methods\\ text\\ space\\ combined\\ averaging\\ area\\ apply\\ recognition\\ bit\end{tabular}} & \multirow{10}{*}{\begin{tabular}[c]{@{}c@{}}paths\\ close\\ path\\ make\\ numbering\\ channels\\ rep\\ scalars\\ anism\\ viously\end{tabular}} & \multirow{10}{*}{\begin{tabular}[c]{@{}c@{}}total\\ paths\\ global\\ path\\ time\\ fixed\\ function\\ yields\\ close\\ computation\end{tabular}} \\
                                                                                                                                                               &                                                                                                                                                                   &                                                                                                                                                                 &                                                                                                                                                  &                                                                                                                                                            &                                                                                                                                                       &                                                                                                                                                &                                                                                                                                                  \\
                                                                                                                                                               &                                                                                                                                                                   &                                                                                                                                                                 &                                                                                                                                                  &                                                                                                                                                            &                                                                                                                                                       &                                                                                                                                                &                                                                                                                                                  \\
                                                                                                                                                               &                                                                                                                                                                   &                                                                                                                                                                 &                                                                                                                                                  &                                                                                                                                                            &                                                                                                                                                       &                                                                                                                                                &                                                                                                                                                  \\
                                                                                                                                                               &                                                                                                                                                                   &                                                                                                                                                                 &                                                                                                                                                  &                                                                                                                                                            &                                                                                                                                                       &                                                                                                                                                &                                                                                                                                                  \\
                                                                                                                                                               &                                                                                                                                                                   &                                                                                                                                                                 &                                                                                                                                                  &                                                                                                                                                            &                                                                                                                                                       &                                                                                                                                                &                                                                                                                                                  \\
                                                                                                                                                               &                                                                                                                                                                   &                                                                                                                                                                 &                                                                                                                                                  &                                                                                                                                                            &                                                                                                                                                       &                                                                                                                                                &                                                                                                                                                  \\
                                                                                                                                                               &                                                                                                                                                                   &                                                                                                                                                                 &                                                                                                                                                  &                                                                                                                                                            &                                                                                                                                                       &                                                                                                                                                &                                                                                                                                                  \\
                                                                                                                                                               &                                                                                                                                                                   &                                                                                                                                                                 &                                                                                                                                                  &                                                                                                                                                            &                                                                                                                                                       &                                                                                                                                                &                                                                                                                                                  \\
                                                                                                                                                               &                                                                                                                                                                   &                                                                                                                                                                 &                                                                                                                                                  &                                                                                                                                                            &                                                                                                                                                       &                                                                                                                                                &                                                                                                                                                  \\ \hline
\end{tabular}
}
\caption{\small The worst four topics produced by LDA, in terms of per-topic coherence score, and their corresponding NEA topics, with LDA trained on the \emph{NIPS} corpus for $K$=$7,000$.}
\label{fig:worst-top-topic}
\end{figure*}
\begin{figure*}[t]
\centering
\resizebox{0.9\textwidth}{!}{
\begin{tabular}{|c|c|c|c|c|c|c|c|}
\hline
LDA                                                                                                                                             & NEA                                                                                                                                                             & LDA                                                                                                                                      & NEA                                                                                                                                                      & LDA                                                                                                                                       & NEA                                                                                                                                                             & LDA                                                                                                                                       & NEA                                                                                                                                                   \\ \hline
\multirow{10}{*}{\begin{tabular}[c]{@{}c@{}}share\\ pittsburgh\\ aa\\ aaa\\ ab\\ abandon\\ abandoned\\ abc\\ abdul\\ aberrational\end{tabular}} & \multirow{10}{*}{\begin{tabular}[c]{@{}c@{}}International\\ common\\ share\\ pittsburgh\\ general\\ agreement\\ tender\\ market\\ june\\ dividend\end{tabular}} & \multirow{10}{*}{\begin{tabular}[c]{@{}c@{}}tonnes\\ yr\\ aa\\ aaa\\ ab\\ abandon\\ abandoned\\ abc\\ abdul\\ aberrational\end{tabular}} & \multirow{10}{*}{\begin{tabular}[c]{@{}c@{}}announced\\ tonnes\\ addition\\ asked\\ accounts\\ shares\\ surplus\\ secretary\\ heavy\\ held\end{tabular}} & \multirow{10}{*}{\begin{tabular}[c]{@{}c@{}}blah\\ aa\\ aaa\\ ab\\ abandon\\ abandoned\\ abc\\ abdul\\ aberrational\\ abide\end{tabular}} & \multirow{10}{*}{\begin{tabular}[c]{@{}c@{}}blah\\ company\\ account\\ advantage\\ acquisitions\\ loss\\ proposed\\ considered\\ announced\\ base\end{tabular}} & \multirow{10}{*}{\begin{tabular}[c]{@{}c@{}}dlrs\\ aa\\ aaa\\ ab\\ abandon\\ abandoned\\ abc\\ abdul\\ aberrational\\ abide\end{tabular}} & \multirow{10}{*}{\begin{tabular}[c]{@{}c@{}}debt\\ canadian\\ today\\ canada\\ decline\\ competitive\\ conditions\\ dlrs\\ price\\ week\end{tabular}} \\
                                                                                                                                                &                                                                                                                                                                 &                                                                                                                                          &                                                                                                                                                          &                                                                                                                                           &                                                                                                                                                                 &                                                                                                                                           &                                                                                                                                                       \\
                                                                                                                                                &                                                                                                                                                                 &                                                                                                                                          &                                                                                                                                                          &                                                                                                                                           &                                                                                                                                                                 &                                                                                                                                           &                                                                                                                                                       \\
                                                                                                                                                &                                                                                                                                                                 &                                                                                                                                          &                                                                                                                                                          &                                                                                                                                           &                                                                                                                                                                 &                                                                                                                                           &                                                                                                                                                       \\
                                                                                                                                                &                                                                                                                                                                 &                                                                                                                                          &                                                                                                                                                          &                                                                                                                                           &                                                                                                                                                                 &                                                                                                                                           &                                                                                                                                                       \\
                                                                                                                                                &                                                                                                                                                                 &                                                                                                                                          &                                                                                                                                                          &                                                                                                                                           &                                                                                                                                                                 &                                                                                                                                           &                                                                                                                                                       \\
                                                                                                                                                &                                                                                                                                                                 &                                                                                                                                          &                                                                                                                                                          &                                                                                                                                           &                                                                                                                                                                 &                                                                                                                                           &                                                                                                                                                       \\
                                                                                                                                                &                                                                                                                                                                 &                                                                                                                                          &                                                                                                                                                          &                                                                                                                                           &                                                                                                                                                                 &                                                                                                                                           &                                                                                                                                                       \\
                                                                                                                                                &                                                                                                                                                                 &                                                                                                                                          &                                                                                                                                                          &                                                                                                                                           &                                                                                                                                                                 &                                                                                                                                           &                                                                                                                                                       \\
                                                                                                                                                &                                                                                                                                                                 &                                                                                                                                          &                                                                                                                                                          &                                                                                                                                           &                                                                                                                                                                 &                                                                                                                                           &                                                                                                                                                       \\ \hline
\end{tabular}
}
\caption{\small The four topics that were most improved by NEA over the original LDA topic, in terms of the difference between per-topic coherence score, with LDA trained on the \emph{Reuters-150} corpus for $K$=$7,000$.}
\label{fig:diff-worst-top-topic}
\end{figure*}

\subsection{Training NEA for LDA }
\label{NEA}
To train the NEA reconstruction of LDA, we start with pre-trained LDA parameters: document-topic distributions $\boldsymbol{\Theta}_{LDA}$, topic-word distributions $\boldsymbol{\Phi}_{LDA}$, and topic assignments $Z$.  
%
Given the input LDA (or other) topic model, our ideal objective function to train NEA is $D_{KL}(p_{LDA}||p_{NEA})$.  It can be seen that minimizing $D_{KL}(p_{LDA}||p_{NEA})$ is equivalent to maximizing $E_{p_{LDA}(w,z)}[p(w,z;\mathbf{V})]$. This suggests a procedure where minibatches are drawn from the topic model, and are used to update the parameters $\mathbf{V} = \{\mathbf{V}^{(W)\prime}, \bar{\mathbf{V}}^{(K)}\}$ via stochastic gradient descent.  We construct minibatches of input topics $z$ and target words $w$ 
by repeatedly drawing a document index $d$ uniformly at random, drawing a topic $z$ from that document's $\boldsymbol{\Theta}_{LDA}^{(d)}$ and sampling a word $w$ from drawn topic $\boldsymbol{\Phi}_{LDA}^{(z_{d})}$. 
%
Then, we would take a gradient step on
 $\log p(w,z|\textbf{V},b,\boldsymbol{\Theta}_{LDA}) = \log p(w|z,\textbf{V},b) + \mbox{const}$ to update $\textbf{V}$.
%
However, as for other embedding models, normalization over the dictionary becomes a bottleneck in the stochastic gradient updates.  
%
Since noise-contrastive estimation (NCE)~\cite{mnih2013learning,gutmann2010noise,gutmann2012noise} has been shown to be an asymptotically consistent estimator of the MLE in the number of noise samples~\cite{gutmann2012noise}, it is a principled approximation of our $E_{p_{LDA}(\mbox{data})}[p(\mbox{data};\mathbf{V})]$ objective. In practice, however, we obtained better performance using negative sampling (NEG) \cite{mikolov2013distributed}, which further approximates the NCE objective as
\begin{equation*}
    \log \sigma({v'_{w}}^\intercal \bar{v}_{z}) + \sum_{i=1}^k E_{w_i \sim p_n(w)}{\log \sigma(-{v'_{w_i}}^\intercal \bar{v}_{z}))} \mbox{ ,}
\end{equation*}
where $p_n(w)$ is a ``noise'' distribution, and $k$ is the number of ``negative'' samples drawn from it per word.  
Having learned the embeddings, we recover NEA's ``smoothed'' encodings of the topics:
\begin{equation}
   \boldsymbol{\Phi}_{NEA} \propto exp(\textbf{V}^{(W)\prime \intercal}\boldsymbol{\bar{V}}^{(K)}) \mbox{ .}  
\end{equation}
Finally, we construct document vectors by summing the corresponding (normalized) topic vectors according to the pre-trained LDA model's topic assignments $Z$, for each token of that document. We normalize all document vectors to unit length to avoid any impact of the length of the document on the scale of the features, to produce the final document embeddings $\mathbf{V}^{(D)}$. 
%
%
%
The pseudocode for training NEA to mimic LDA is shown in Algorithm~\ref{alg-main}.

\subsection{General NEA Algorithm}
More generally, the NEA method can be extended to encode any topic model's parameters, which are typically conditional distributions given a single parent assignment, $P(a_{i}|parent(a_{i}))$, into \emph{vector representations} $\mathbf{V}^{(i)}$, $\mathbf{V}^{(i)\prime}$ while also providing \emph{smoothed versions} of the parameters $P_{NEA}(a_{i}|parent(a_{i}))$. The general learning algorithm of our proposed NEA model for general topic models is shown in Algorithm~\ref{alg-general-nea}. In the embedding steps, for each iteration, we draw samples $a_i$ from the conditional discrete distributions for documents, authors, topics, words, etc., followed by updating the input and output vectors by optimizing log-bilinear classification problems using negative sampling (discussed in Section~\ref{NEA}). In the smoothing steps, we can recover the smoothed version of the parameters $P_{NEA}(a_{i}|parents(a_{i}))$ by the dot product of the corresponding input and output vectors learned in embeddings steps followed by a softmax projection onto the simplex. 
\begin{figure*}[t] 
		\centerline{\includegraphics[width=0.98\textwidth]{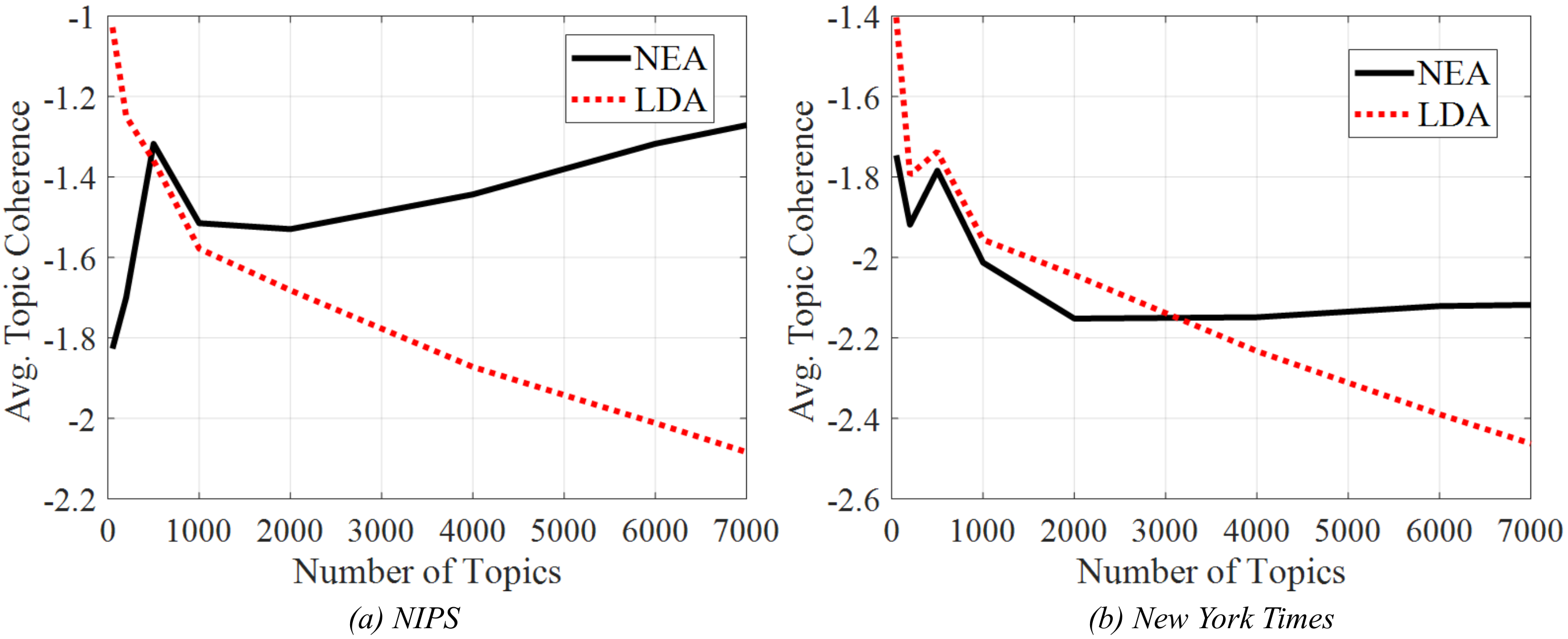}}
		\caption{\small Comparison of average topic coherence $vs.$ number of topics $K$ on four different corpora: (a) \emph{NIPS}, and (b) \emph{New York Times}. NEA generated topics outperform LDA topics in terms of higher average topic coherence when $K$ is large.}
		\label{fig:coherence} 
\end{figure*}
\section{Experiments}
The goals of our experiments were to evaluate the NEA model both as a topic model and as a feature engineering method for classification tasks. We will release the source code of our implementation once the paper is accepted. 

For several experiments, we considered five datasets. First, we use the \emph{NIPS} corpus with $1,740$ scientific articles from years $1987$-$1999$ with $2.3$M tokens, which contains a dictionary size of $13,649$ words. The second dataset contains $4,676$ articles published by the \emph{New York Times}  with a dictionary size of $12,042$ words. We also used another dataset, \emph{Bibtex},\footnote{\url{http://mulan.sourceforge.net/datasets-mlc.html}.} which contains $7,395$ references as documents with a dictionary size of $1,643$ words. Finally, the \emph{Reuters$-$150} news wire articles corpus ($15,500$ articles with dictionary size of $8,349$ words) and \emph{Ohsumed} medical abstracts ($20,000$ articles where classes are $23$ cardiovascular diseases) were used. 

\subsection{Performance for LDA}
\begin{figure*}[t] 
		\centerline{\includegraphics[width=0.9\textwidth]{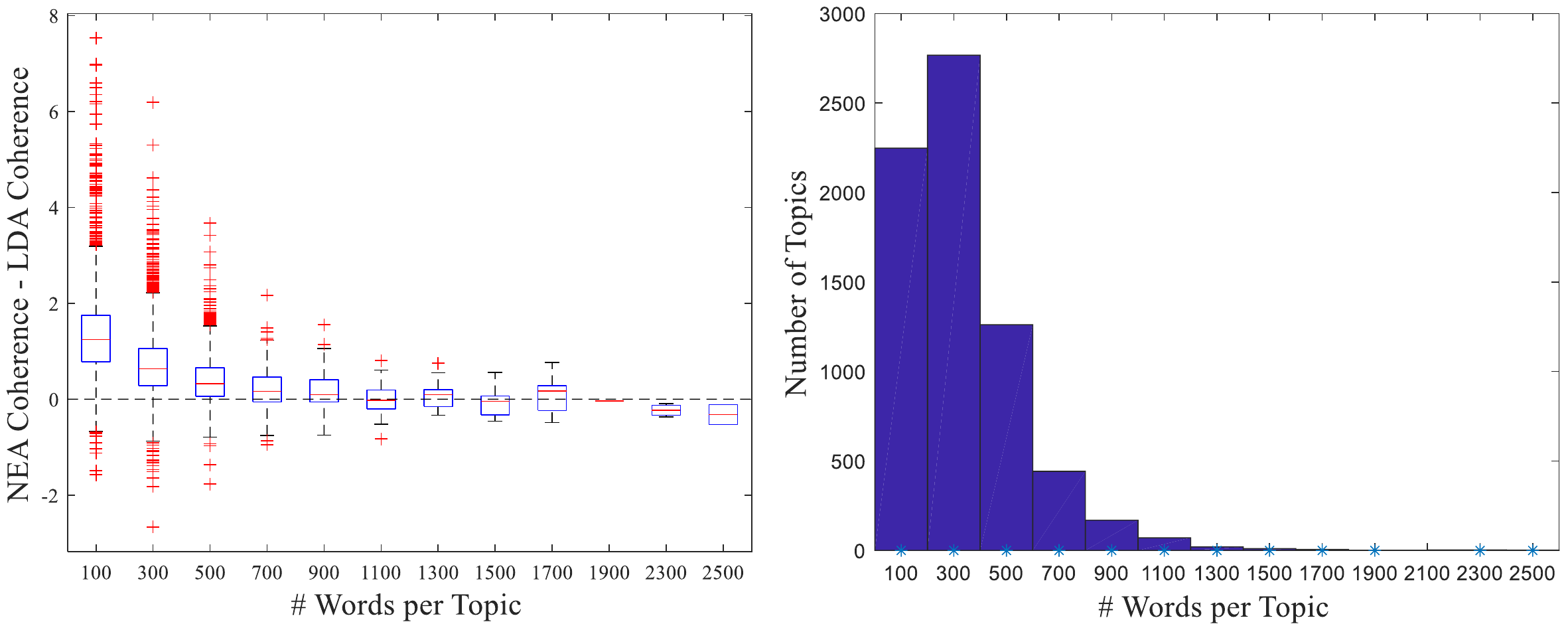}}
		\caption{\small Improvement in coherence of NEA over LDA \emph{vs.} number of words in the topic, $K=7,000$, \emph{NIPS} dataset. Boxplot (left) shows coherence improvement for number of words per topic while histogram (right) shows number of topics in each bin.}
		\label{fig:coherenceDiff} 
\end{figure*}
\begin{figure}[t] 
		\centerline{\includegraphics[width=0.5\textwidth]{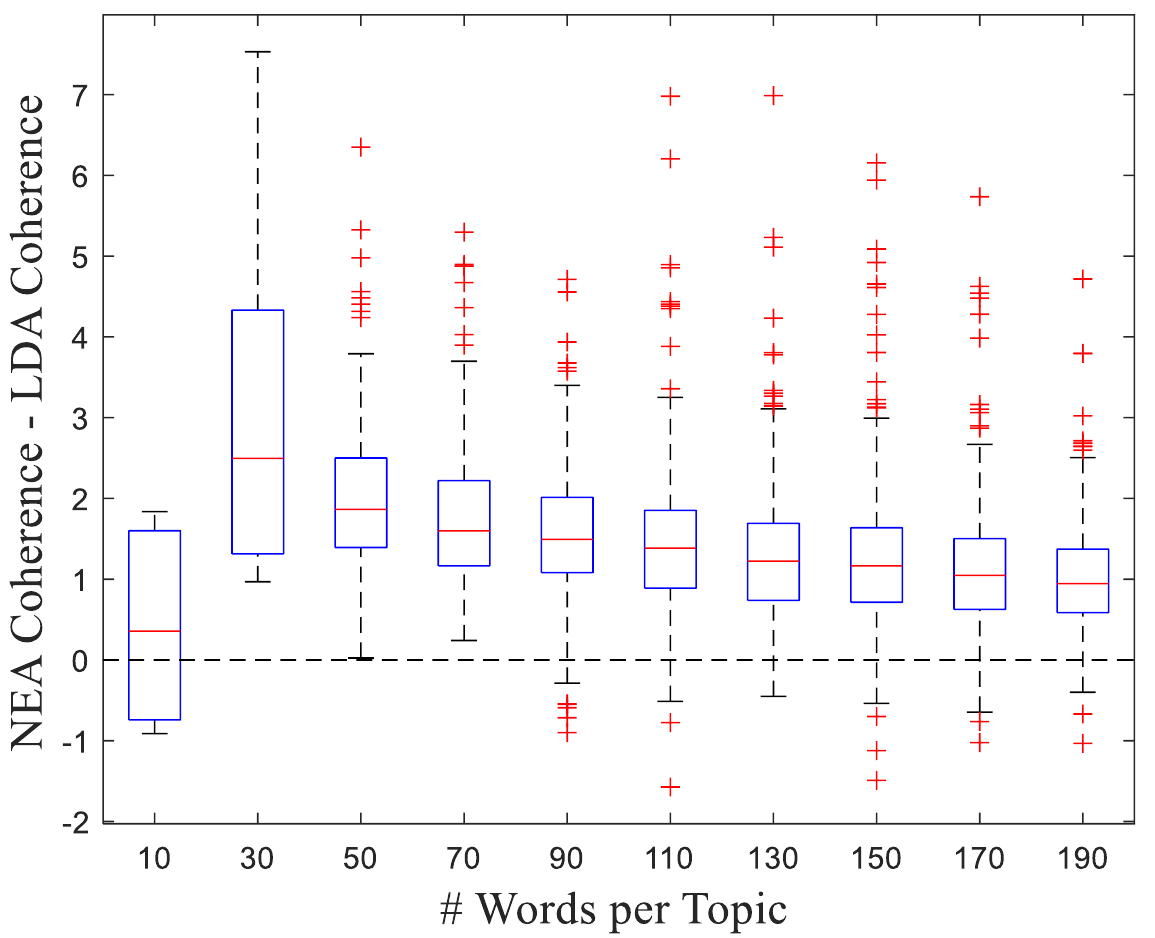}}
		\caption{\small Improvement of bad topics (less than $200$ assigned words) by NEA \emph{vs.} number of words in the topic, $K=7,000$, for \emph{NIPS}.}
		\label{fig:nipsVSnytime} 
\end{figure}
We start our analysis by evaluating how NEA performs at mimicking LDA in terms of topic and embeddings quality.
\subsubsection{Quality of Topics}
 To perform this experiment, we compare the quality of generated topics from LDA and NEA by investigating both qualitative and quantitative results on several data sets.


We fix LDA's hyperparameters at $\alpha$=$0.1$ and $\beta$=$0.01$ when $K$$<$$500$, 
otherwise we use $\alpha$=$0.01$ and $\beta$=$0.001$. LDA was trained using the Metropolis-Hastings-Walker algorithm \cite{li2014reducing}, due to its scalability in the number of topics $K$. In NEA, negative sampling (NEG) was performed for $1$ million minibatches of size $16$ with $300$-dimensional embeddings. In the experiments, we found that NEA generally recovers the same top words for LDA's ``good topics" (example topics are shown in the Appendix). 

To get a quantitative comparison, we compared the topics' UMass coherence metric, which measures the semantic quality of a topic based on its $T$ most probable words (we choose $T=10$ words), thereby quantifying the user's viewing experience~\cite{mimno2011optimizing}. Larger coherence values indicate greater co-occurrence of the words, hence higher quality topics. In Figure~\ref{fig:coherence}, the average topic coherence of LDA and NEA is shown with respect to the number of topics $K$. LDA works well with small $K$ values, but when $K$ becomes large, NEA outperforms LDA in average topic coherence scores on all datasets (see the Appendix for similar results on two other datasets). 

In Figure \ref{fig:worst-top-topic}, we show the four worst topics from LDA, based on per-topic coherence score, and their corresponding NEA topics, when the model was trained on \emph{NIPS} for $K=7,000$. In this case, NEA generated slightly more meaningful topics than LDA. We also identified the most improved topics based on the difference between per-topic coherence scores. In Figure \ref{fig:diff-worst-top-topic}, we show the 4 topics with the largest improvement in coherence scores by NEA, for \emph{Reuters$-$150} with $7,000$ topics. We observe that these LDA topics were uninterpretable, and likely had very few words assigned to them. NEA tends to improve the quality of these ``bad'' topics, e.g. by replacing stop words (or words at the top of the dictionary) with more semantically related ones.
In particular, we found that NEA gave the most improvement for topics with few words assigned to them (see Figure~\ref{fig:coherenceDiff} (left)) and when $K$ becomes large, the majority of topics have few assigned words (see Figure~\ref{fig:coherenceDiff} (right)). As a result, NEA improves the quality of most of the topics. In Figure~\ref{fig:nipsVSnytime}, we showcase the improvement for ``bad topics,'' those which have less than $200$ words assigned to them, by our proposed NEA model on the \emph{NIPS} corpus.

\begin{table*}[t]
\centering
\resizebox{0.88\textwidth}{!}{
\begin{tabular}{cccccccccl}
\hline
Datasets      & \#Classes & \#Topics & Doc2Vec & LDA  & NEA & SG & SG+NEA & Tf-idf\\ \hline
Reuters-150   & 116       & 500 & 55.89  & 64.26   &  67.15 & 70.80 & 72.29 & 73.00 \\
Ohsumed       & 23        & 500    & 34.02     & 32.05 & 34.38  & 37.26  & 38.88 & 43.07 \\ \hline
\end{tabular}
}
\caption{Comparing NEA in document categorization tasks with other baseline methods. Classification accuracy is shown for two different corpora: \emph{Reuters$-$150}, and \emph{Ohsumed}.}
\label{table:doc_cat_1}
\end{table*}
\begin{table*}[]
\centering
\resizebox{1.0\textwidth}{!}{
\begin{tabular}{cccccccc}
\hline
Datasets      & \#Classes & \#Topics & Tf-idf & Tf-idf+LDA & Tf-idf+SG & Tf-idf+NEA & Tf-idf+SG+NEA \\ \hline
Reuters-150   & 116       & 500      & 73.00  & 73.01      & 72.99     & 73.14      & 73.09         \\
Ohsumed       & 23        & 500      & 43.07  & 43.05      & 43.04     & 43.11      & 43.08         \\ \hline
\end{tabular}
}
\caption{Comparing NEA in document categorization tasks along with tf-idf. Classification accuracy is shown for two different corpus: \emph{Reuters$-$150}, and \emph{Ohsumed}. Tf-idf+NEA had the best classification accuracy.}
\label{table:doc_cat_2}
\end{table*}
\begin{figure}[t] 
		\centerline{\includegraphics[width=0.9\columnwidth]{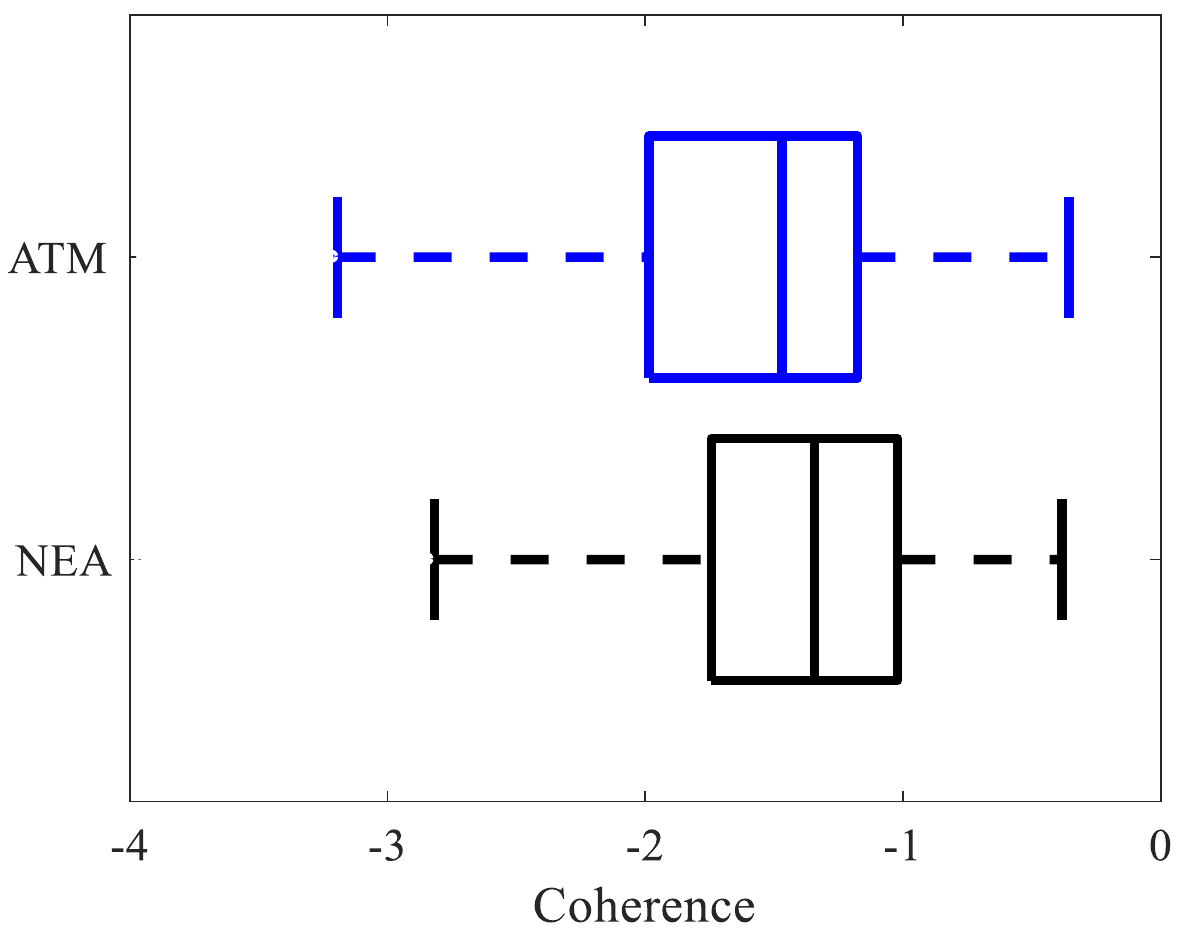}}
		\caption{\small NEA-smoothed ATM outperforms ATM in terms of per topic coherence for $K=1,000$ topics on \emph{NIPS} corpus.}
		\label{fig:ATMvsNEA} 
\end{figure}
\subsubsection{Document Categorization}
In this set of experiments, we tested the performance of the learned vectors using NEA's document embeddings $\textbf{V}^{(D)}$ as features for document categorization/classification. The results are given in Table~\ref{table:doc_cat_1}. We used two standard benchmark datasets: \emph{Reuters$-$150}, and \emph{Ohsumed}.\footnote{All document categorization datasets were obtained from \url{http://disi.unitn.it/moschitti/corpora.htm} .} We used the standard train/test splits from the literature (e.g. for
Ohsumed, 50\% of documents were assigned to training
and to test sets). We also considered tf-idf as a baseline. Logistic regression classifiers were trained on the features extracted on the training set for each method while classification accuracy was computed on the held-out test data. Finally, we compared NEA with LDA as well as several state-of-the-art models such as the skip-gram (SG)~\cite{mikolov2013efficient,mikolov2013distributed}, and paragraph vector (doc2Vec)~\cite{le2014distributed}.

From the results in Table \ref{table:doc_cat_1}, we found that NEA has better accuracy in classification performance than LDA and doc2Vec.
In NEA, the document vectors are encoded at the topic level rather than the word level, so it loses word level information in the embeddings, which turned out to be beneficial for these specific classification tasks, at which SG features outperformed NEA's features. Interestingly, however, when both SG and NEA features were concatenated (SG + NEA), this improved the classification performance over each model's  individual performance. This suggests that the combination of topic-level NEA and word-level SG vectors complement the qualities of each other and both are valuable for performance. Note that the tf-idf baseline, which is notoriously effective for document categorization, outperformed the other features. In Table \ref{table:doc_cat_2}, we show the results when concatenating tf-idf with the other feature vectors from LDA, SG, and NEA, which in many cases improved performance over tf-idf alone. We observed the highest improvement over tf-idf for both document categorization tasks when we concatenated NEA vectors with tf-idf (tf-idf + NEA). This approach outperformed all other feature combinations.  This may be because the topical information in NEA features is complementary to tf-idf, while SG's word-based features are redundant.
\subsection{Performance for ATM}
In the second phase of our experiments, we trained NEA for the author-topic model (ATM)'s generated parameters, with the same hyperparameters we used in the previous section. Similar to the experiment for LDA, NEA improves topic coherence of the ATM generated topics when $K$ is large. Figure~\ref{fig:ATMvsNEA} shows NEA outperforms ATM in terms of per-topic coherence for \emph{NIPS} when $K=1000$.   

We also studied the performance of NEA for smoothing the author-topic distributions. The ATM could be used for a variety of applications such as automated reviewer recommendations~\cite{rosen2004author}, which could benefit from NEA smoothing. Since these applications are based on searching for similar authors, we can treat them as a ranking problem. Following  ~\cite{rosen2004author}, we rank based on the \emph{symmetric KL-divergence} between authors $i$ and $j$:
\begin{equation}
     sKL(i,j) = \sum_{t=1}^{K} [\theta_{it}\log\frac{\theta_{it}}{\theta_{jt}} + \theta_{jt}\log\frac{\theta_{jt}}{\theta_{it}}] \mbox{,}
\end{equation}
where $\theta_i$ is the $i$th author's distribution over topics.  Using this distance metric, we searched for similar authors in the \emph{NIPS} corpus for the $125$ out of $2037$ authors who wrote at least 5 papers. We reported the mean reciprocal rank (MRR) based on the rank of the most similar co-author. Table {\ref{table:MRR_ATM}} shows the improvement in MRR using author vectors generated from NEA over the author-topic parameters of the ATM.  Further improvement was achieved by the NEA-smoothed version of the ATM's parameters which also outperformed author vectors generated from a tf-idf baseline at this task.  

\begin{table}[t]
\resizebox{0.5\textwidth}{!}{
\begin{tabular}{lcccc}
\hline
     & ATM & NEA embeddings & Tf-idf & NEA smoothing \\ \hline
NIPS & 0.016                  & 0.018  & 0.019             & 0.021                           \\ \hline
\end{tabular}
}
\caption{Mean reciprocal rank for co-author retrieval.}
\label{table:MRR_ATM}
\end{table}
\begin{figure}[t] 
\vspace{-0.2cm}
		\centerline{\includegraphics[width=1.05\columnwidth]{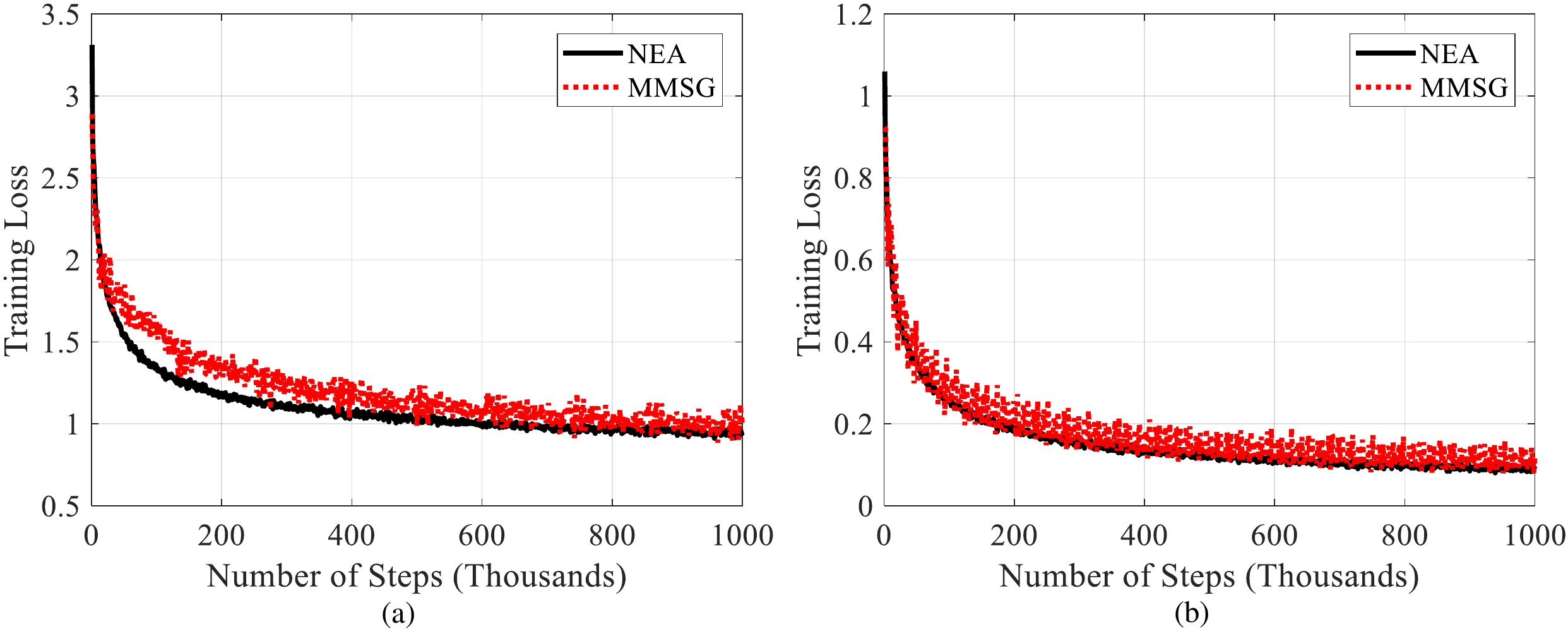}}
		\caption{\small Loss curve of NEA and MMSG training for MMSGTM when $K=1,000$ topics on (a) \emph{NIPS} and (b) \emph{Reuters-150} corpus.}
		\label{fig:training_loss} 
\vspace{-0.4cm}
\end{figure}
\subsection{Performance for MMSGTM}
Finally, we train NEA to reparameterize the mixed membership skip-gram topic model (MMSGTM)~\citep{foulds2018mixed}. We used the same hyperparameter values as in previous experiments, while setting MMSGTM-specific hyperparameters to the values suggested in ~\cite{foulds2018mixed}. The original MMSG algorithm learns topic embeddings based on the MMSGTM's cluster assignments $Z$, while NEA uses simulated data from the topic model.  The NEA method is arguably a more principled method to accomplish the embedding as it has an explicit objective function.  We found that NEA smooths and slightly improves the speed of the training process (shown in Figure~\ref{fig:training_loss}), while greatly reducing memory requirements as the topic assignments $Z$ need not be stored.  NEA training for MMSG improves over MMSGTM at classification and performs similarly to ~\citet{foulds2018mixed}'s algorithm (see results in the Appendix).



%
\section{Conclusion}
We have proposed neural embedding allocation (NEA) for learning interpretable vector-space embeddings of words, documents, topics, and authors by deconstructing topic models to reveal underlying semantic representations.
Our experimental results show that our proposed NEA method successfully mimics topic models with nuanced vector representations, while performing better than them at many tasks. The proposed NEA algorithm can smooth out topic models' parameters to improve topic coherence and author modeling, and produces vector representations which improve document categorization performance. We plan to use NEA to study and address gender bias issues in natural language processing.   

\bibliography{ref}
\bibliographystyle{named}



\newpage
\appendix

\section{Background and Related Work}
For completeness, and to establish notation, we provide background on topic models and word embeddings.

\subsection{Latent Dirichlet Allocation}
Probabilistic topic models, for example, LDA~\cite{blei2003latent} use latent variables to encode co-occurrences between words in text corpora and other bag-of-words represented data. A simple way to model text corpora is using multinomial naive Bayes with a latent cluster assignment for each document, which is a multinomial distribution over words, called a \emph{topic} $k \in \{1, ... K\}$. LDA topic models improve over naive Bayes using mixed membership, by relaxing the condition that all words in a document $d$ belong to the same topic. In LDA's generative process, for each word $w_{di}$ of a document $d$, a topic assignment $z_{di}$ is sampled from document-topic distribution $\theta^{(d)}$ followed by drawing the word from topic-word distribution $\phi^{(z_{di})}$ (see Table 1 in the main paper, bottom-right). Dirichlet priors encoded by $\alpha_{k}$ and $\beta_{w}$ are used for these parameters, respectively.

\subsection{Author Topic Model}
Author-topic model (ATM) is a probabilistic model for both author and topics by extending LDA to include authorship information \cite{rosen2004author}. In the generative process of ATM, for each word $w_{di}$ of a document $d$, an author assignment $a_{di}$ is uniformly chosen from number of authors $A_{d}$ and then a topic assignment $z_{di}$ is sampled from author-topic distribution $\theta^{(a_{di})}$ followed by drawing the word from topic-word distribution $\phi^{(z_{di})}$ as follows:
\begin{itemize}
\item	For each document  $d$
\vspace{-0.2cm}
\begin{itemize}
    \item For each word in the document $w_{di}$
    \begin{itemize}
        \item Draw $a_{di} \sim \mbox{Uniform}(\frac{1}{|A_{d}|})$
        \item Draw $z_{di} \sim \mbox{Discrete}(\theta^{(a_{di})})$
        \item Draw $w_{di}  \sim \mbox{Discrete}(\phi^{(z_{di})})$
    \end{itemize}
\end{itemize}
\end{itemize}

Like LDA, similar Dirichlet priors $\alpha_{a}$ and $\beta_{w}$ are used for $\theta^{(a)}$ and $\phi^{(z)}$ parameters, respectively.

\subsection{MMSG Topic Model}
To show the generality of our approach to topic models we also consider our method to a recent model called the mixed membership skip-gram topic model (MMSGTM) \cite{foulds2018mixed}, which combines ideas from topic models and word embeddings to recover domain specific embeddings for small data.
The generative model for MMSGTM is:
\begin{itemize}
\item For each word $w_{i}$ in the corpus
\begin{itemize}
    \item Sample a topic $z_{i} \sim \mbox{Discrete}(\theta^{w_{i}})$
    \item For each word $w_{c} \in context(i)$
    \begin{itemize}
        \item Sample a context word\\ $w_{c}  \sim \mbox{Discrete}(\phi^{z_{i}})$ .
    \end{itemize}
\end{itemize}
\end{itemize}
Finally, the mixed membership skip-gram model (MMSG) is trained for word and topic embeddings with the topic assignments $z$ as input and surrounding $w_{c}$ as output. Since MMSG training depends on the topic assignments as well as the whole corpus, it is not scalable for big data.
\begin{figure*}[t]
\centering
\resizebox{1.0\textwidth}{!}{
\begin{tabular}{|c|c|c|c|c|c|c|c|}
\hline
LDA                                                                                                                                    & NEA                                                                                                                                        & LDA                                                                                                                                       & NEA                                                                                                                                          & LDA                                                                                                                                                     & NEA                                                                                                                                               & LDA                                                                                                                                                           & NEA                                                                                                                                                   \\ \hline
\begin{tabular}[c]{@{}c@{}}bayesian\\ prior\\ bayes\\ posterior\\ framework\\ priors\\ likelihood\\ bars\\ note\\ compute\end{tabular} & \begin{tabular}[c]{@{}c@{}}bayesian\\ bayes\\ posterior\\ priors\\ likelihood\\ prior\\ framework\\ note\\ probability\\ bars\end{tabular} & \begin{tabular}[c]{@{}c@{}}images\\ image\\ recognition\\ vision\\ pixel\\ techniques\\ pixels\\ visual\\ computed\\ applied\end{tabular} & \begin{tabular}[c]{@{}c@{}}images\\ image\\ visual\\ recognition\\ pixels\\ pixel\\ illumination\\ intensity\\ pairs\\ matching\end{tabular} & \begin{tabular}[c]{@{}c@{}}phrase\\ sentences\\ clause\\ structure\\ sentence\\ phrases\\ syntactic\\ connectionist\\ tolerance\\ previous\end{tabular} & \begin{tabular}[c]{@{}c@{}}sentences\\ phrase\\ structure\\ sentence\\ clause\\ activation\\ connectionist\\ phrases\\ roles\\ agent\end{tabular} & \begin{tabular}[c]{@{}c@{}}regression\\ linear\\ ridge\\ quadratic\\ squared\\ nonparametric\\ dimensionality\\ variables\\ smoothing\\ friedman\end{tabular} & \begin{tabular}[c]{@{}c@{}}regression\\ linear\\ ridge\\ quadratic\\ variables\\ nonparametric\\ squared\\ multivariate\\ kernel\\ basis\end{tabular} \\ \hline
\end{tabular}
}
\caption{\small Randomly selected topic pairs from LDA and NEA, with LDA trained on the \emph{NIPS} corpus for $K$=$2,000$.}
\label{fig:top-topic}
\end{figure*}

\begin{figure*}[t]
\centering
\resizebox{1.0\textwidth}{!}{
\begin{tabular}{|c|c|c|c|c|c|c|c|}
\hline
LDA           & NEA         & LDA          & NEA          & LDA           & NEA           & LDA      & NEA          \\ \hline
learning      & learning    & blake        & models       & insertion     & space         & strain   & structure    \\
steps         & steps       & condensation & exp          & hole          & reinforcement & mars     & length       \\
computer      & computer    & isard        & blake        & gullapalli    & learning      & yield    & variance     \\
testing       & testing     & models       & similar      & reinforcement & fig           & rolling  & equal        \\
observation   & people      & observations & condensation & smoothed      & insertion     & mill     & mars         \\
predetermined & bin         & entire       & modified     & reactive      & hole          & cart     & strain       \\
cheng         & observation & oxford       & generally    & extreme       & fit           & tuning   & weight       \\
utilizes      & efficient   & rabiner      & cortical     & ram           & gullapalli    & material & intelligence \\
efficient     & utilizes    & gelb         & isard        & gordon        & maximum       & friedman & cycle        \\
updating      & birth       & north        & consisting   & consecutive   & regions       & plot     & friedman     \\ \hline
\end{tabular}
}
\caption{\small The worst four topics produced by NEA, in terms of per-topic coherence score, and their corresponding LDA topics, with LDA trained on the \emph{NIPS} corpus for $K$=$7,000$..}
\label{fig:Worst-top-topic_By_NEA}
\end{figure*}

\begin{figure*}[t] 
		\centerline{\includegraphics[width=0.98\textwidth]{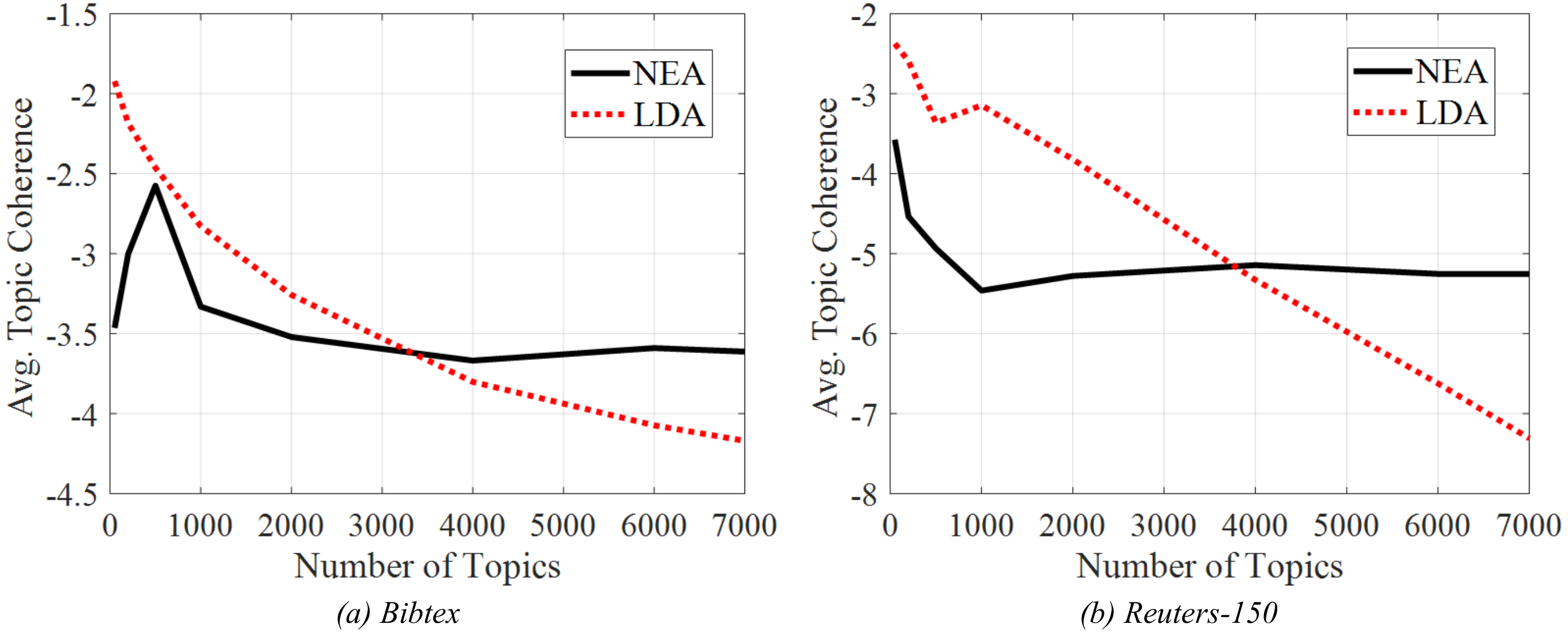}}
		\caption{\small Comparison of average topic coherence $vs.$ number of topics $K$ on four different corpora: (a) \emph{Bibtex}, and (b) \emph{Reuters-150}. NEA generated topics outperform LDA topics in terms of higher average topic coherence when $K$ is large}
		\label{fig:otherCoherence} 
\end{figure*}
\begin{figure}[t] 
		\centerline{\includegraphics[width=0.5\textwidth]{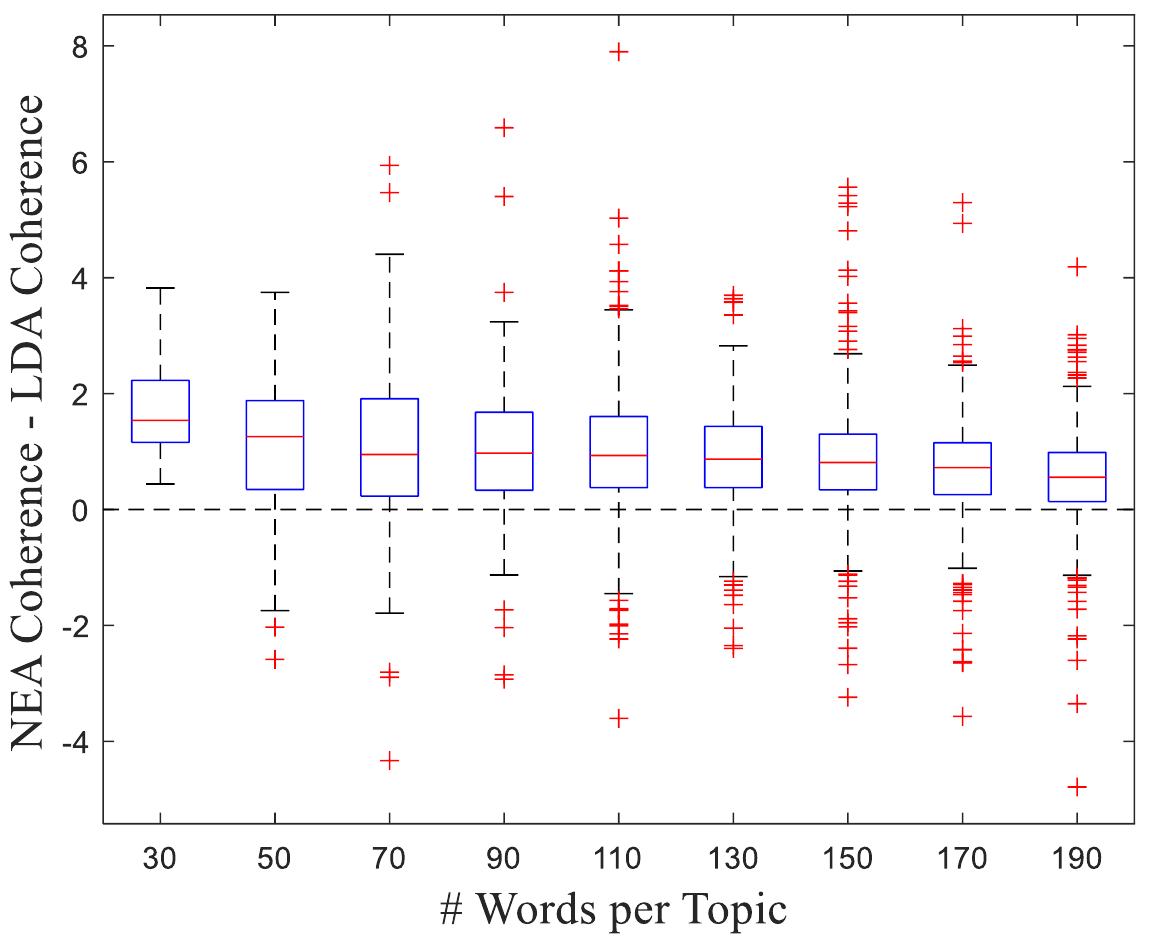}}
		\caption{\small Improvement of bad topics (less than $200$ assigned words) by NEA \emph{vs.} number of words in the topic, $K=7,000$, for \emph{NYTime} corpus.}
		\label{fig:nytime} 
\end{figure}

\subsection{Word Embeddings}
Traditional probabilistic language models predict words given their context words using a joint probability for sequences of words in a language~\cite{bengio2003neural} based on distributed representations~\cite{hinton1986learning} from neural network weights. Later, word embeddings were found to be useful for semantic representations of words, even without learning a full joint probabilistic language model. In particular, the skip-gram model is an effective method for learning better quality vector representations of words from big unstructured text data.   

The skip-gram~\cite{mikolov2013distributed} is a log-bilinear classifier for predicting words that occur in the context of other words in a document, where the context is typically defined to be a small window around the word. For a sequence of input training words, the objective of the skip-gram model is to maximizing the average log probability of the output context words given the input word. We can think of it as a certain parameterization of a set of discrete distributions, $p(w_c|w_i)$, where $w_c$ is a context word and $w_i$ is an ``input'' word, and both $w_c$ and $w_i$ range over the $W$ words in the dictionary (see Table 1 in the main paper, top-left).  In the simplest case, these discrete distributions have the form:
\begin{equation}\label{eq:skip-gram}
    p(w_c|w_i) \propto exp({v'_{w_c}}^\intercal v_{w_i}) \mbox{ .} 
\end{equation}
where, $v'_{w_c}$ and $v_{w_i}$ are vector embeddings of context words and input words, respectively, with dimensionality $V$. 
\section{Additional Experiments}
In this section, we demonstrate our results on other datasets by repeating the similar experiments on them. First, we found that most of the topics produced by both models are interpretable, and NEA was able to approximately recover the original LDA's topics. In Figure~\ref{fig:top-topic}, we show a few randomly selected example topics, where LDA was trained on the NIPS corpus for $K=2,000$. 

In the main paper, we show four worst topics from LDA with their corresponding NEA topics. Here in Figure \ref{fig:Worst-top-topic_By_NEA}, we show the four worst topics generated from NEA, based on per-topic coherence score, and their corresponding LDA generated topics for the same model. In this case, LDA generates slightly more meaningful topics than NEA.

We showed previously that NEA improves LDA topics in terms of average coherence for \emph{NIPS}, and \emph{NYTime} when $K$ is large. We repeated the same experiment for \emph{Bibtex}, and \emph{Reuters-150} which also gave the same trend in average coherence result (see Figure~\ref{fig:otherCoherence}).  

We showcase again the improvement of ``bad topics'' those which have less than $200$ words assigned to them, by the NEA model for \emph{NYTime} corpus in Figure~\ref{fig:nytime}.
\begin{table}[t]
\resizebox{0.45\textwidth}{!}{
\begin{tabular}{lccc}
\hline
             & MMSGTM & NEA & MMSG     \\ \hline
Reuters-150 & 66.97 & 68.14  & 68.26  \\
Ohsumed      & 32.41 & 34.42 & 34.78  \\ \hline
\end{tabular}
}
\caption{Comparing NEA in document categorization tasks with MMSG, when both models trained for MMSGTM. Classification accuracy is shown for two different corpus: \emph{Reuters$-$150}, and \emph{Ohsumed}.}
\label{table:MMSGvsNEA}
\end{table}

To evaluate the performance of NEA for MMSGTM, we perform document categorization tasks for NEA and MMSG as shown in Table~\ref{table:MMSGvsNEA} when both models trained for MMSGTM. Both NEA and MMSG improves accuracy of this downstream task comparing to MMSGTM for \emph{Reuters$-$150}, and \emph{Ohsumed} dataset while MMSG maintains slightly higher accuracy than NEA.   

\end{document}